\documentclass{article}
\usepackage{hyperref}
\usepackage{booktabs} % For better table rules
\usepackage{array}    % For better column formatting
\usepackage{spconf,amsmath,graphicx}
\hypersetup{hypertex=true,
            colorlinks=true,
            linkcolor=blue,
            anchorcolor=blue,
            citecolor=blue}
\usepackage{graphicx}
\usepackage{amssymb}
\usepackage{caption}
\title{Investigate the Low-level Visual Perception in Vision-Language based Image Quality Assessment 
}
\name{Yuan Li, Zitang Sun, Yen-Ju Chen, Shin'ya Nishida\* \thanks{This work was supported by Spring Fellowship, Grant Number JPMJFS2123; and KAKENHI, Grant Number 24H00721.}}

\address{Graduate School of Informatics, Kyoto University,  Kyoto, {606-8501},  {Japan}}

\begin{document}
%\ninept
%
\maketitle
\begin{abstract}

Recent advances in Image Quality Assessment (IQA) have leveraged Multi-modal Large Language Models (MLLMs) to generate descriptive explanations. However, despite their robust visual perception capabilities, these methods often struggle to reliably detect fundamental low-level visual distortions—such as blur, noise, or compression—and may yield inconsistent evaluations across repeated inferences. This observation raises the question of whether MLLM-based IQA systems truly perceive and recognize essential visual features.
To investigate this issue, we introduce a low-level distortion perception task that requires models to classify specific distortions. Our detailed component-wise analysis reveals that while MLLMs are structurally capable of representing these distortions, they tend to overfit training data templates, thereby introducing biases in quality scoring. As a consequence, critical visual features pertinent to IQA are lost during the vision-language alignment transfer stage.
Moreover, by computing the semantic distance between image features and corresponding semantic tokens before and after component-wise fine-tuning, we demonstrate that enhancing the alignment of the vision encoder substantially improves distortion recognition accuracy—from 14.92\% to 84.43\%. More broadly, these findings suggest that incorporating dedicated training constraints on the vision encoder can foster more text-explainable features, ultimately enabling LLM-based pipelines to generate more coherent and interpretable reasoning in vision-centric tasks.
\end{abstract}
\section{Introduction}
\label{sec:intro}
Recent advancements in large language models (LLMs) have led to their integration into vision tasks, giving rise to  Multi-modal Large Language Models (MLLMs) \cite{flamingo,llava, mplug-owl2}. By combining powerful text-based reasoning with vision encoders, these models have shown promise in addressing various computer vision problems, including Image Quality Assessment (IQA). Compared to earlier approaches, MLLMs offer notable advantages in both visual perception and reasoning, making them increasingly attractive for downstream tasks that require reliable interpretation of visual features \cite{qinstruct,grouding-dino, seagull}. For instance, in blind IQA, contemporary MLLMs not only predict numerical quality scores but also generate descriptive explanations of their reasoning process, highlighting key visual attributes such as brightness, and low-level distortions.
\begin{figure}[t]
  \centering
  \includegraphics[width=0.5\textwidth]{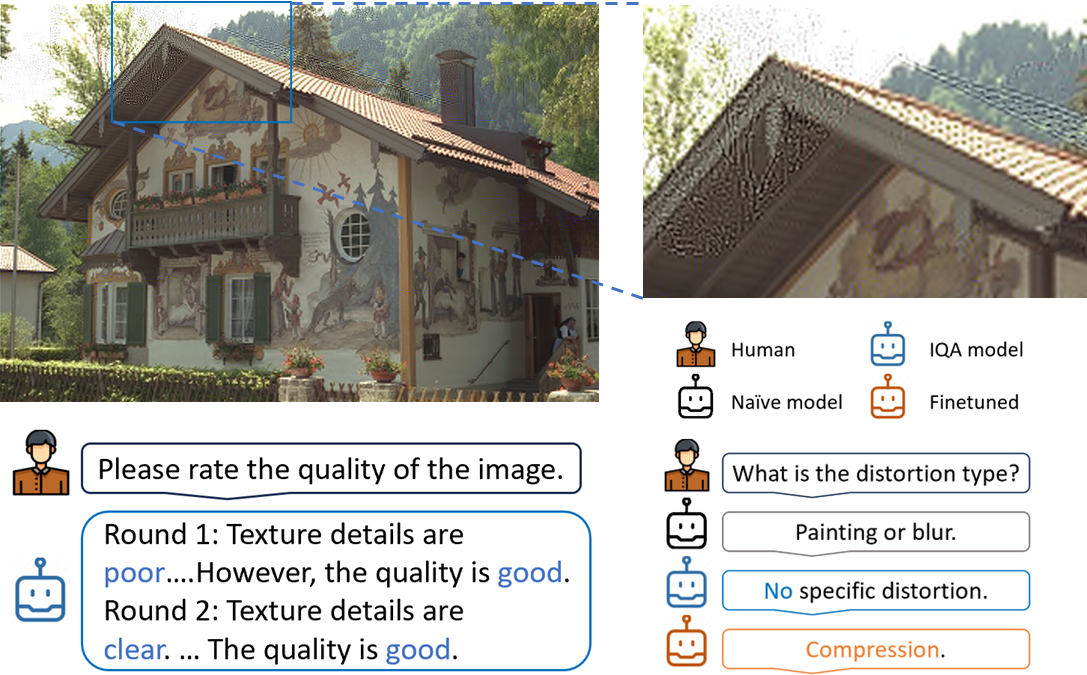}
  \caption{\textbf{Performance on Visual Perception.} The left panel highlights the inconsistency of an IQA model’s perceptions. Meanwhile, the right panel compares how the naive and fine-tuned models classify distortion types. Both models are based on the mPLUG-Owl2 architecture \cite{mplug-owl2}, while the IQA model used for quality assessment is Q-Instruct \cite{qinstruct}. The evaluated sample is a compressed image from the LIVE dataset \cite{live}.}
  \label{fig:f1}
\end{figure}

However, our experiments reveal that certain MLLM-based IQA models \cite{qinstruct, qalign} struggle to recognize fundamental low-level distortions, achieving only 9.05\% accuracy in distortion type prediction (see Table \ref{t1}). As illustrated in Fig. \ref{fig:f1}, an IQA model \cite{qinstruct} inconsistently assesses the quality of an image degraded by compression noise, erroneously labeling it as having “No specific distortion.” These discrepancies raise concerns about the reliability of MLLMs in perception-driven reasoning tasks. A crucial question emerges: Are current MLLM-based IQA models truly capable of perceiving essential visual features? If so, what is the inner mechanism within MLLMs, and how can their perception capabilities be effectively improved?

To address this, we introduce a fundamental visual perception task designed to evaluate the ability of MLLMs to recognize low-level distortions. Specifically, we transform four widely used IQA datasets—LIVE \cite{live}, CSIQ \cite{csiq}, KADID \cite{kadid}, and TID2013 \cite{tid2013}—into a unified multi-modal dataset annotated with seven common distortion categories. We train and evaluate a baseline mPLUG-Owl2 \cite{mplug-owl2} model on this dataset and compare its performance against Q-Instruct \cite{qinstruct}. To gain deeper insights, we decompose the task into two key stages: visual perception and reasoning.

Our findings indicate that while MLLMs are structurally capable of perceiving distortions, they tend to overfit training data templates, introducing biases in quality scoring and ignoring the other fundamental visual features. As a result, critical IQA-relevant visual features are lost during the vision-language alignment stage, and the LLM fails to retrieve visual-feature-related semantic tokens during reasoning. Moreover, the visual perception stage plays a more decisive role in task performance than reasoning. A model with only a fine-tuned vision encoder achieves 88.80\% accuracy, compared to 74.24\% when fine-tuning is applied only at the reasoning stage. This highlights the crucial role of the vision encoder in maintaining feature integrity and alignment.

Furthermore, for MLLMs to achieve robust visual perception, their image representations should be tightly aligned with the semantic space of distortions. To measure this alignment, we compute the semantic distance between image features and distortion tokens before and after fine-tuning. Our results reveal a clear functional division: fine-tuning the vision encoder primarily reduces the semantic gap between visual and linguistic features, while fine-tuning the projection module increases the likelihood of generating the correct distortion token in linguistic output. Therefore, the current observation supports that integrating targeted training constraints on the vision extraction module is critical for bridging the semantic gap between visual features and their corresponding linguistic representations. More broadly, these findings suggest that incorporating dedicated training constraints on the vision encoder can foster the development of more text-explainable features. Such an approach not only improves the accuracy of low-level distortion recognition but also enables LLM-based pipelines to generate more coherent and interpretable reasoning in vision-centric tasks.

In summary, our main contributions are:
\begin{enumerate}
\item \textbf{Unified Distortion Dataset:} We integrate four widely used IQA datasets into a unified multi-modal dataset designed to evaluate fundamental visual feature perception in MLLMs.

\item \textbf{Findings on MLLM Limitations:} Through this dataset, we reveal the inefficiencies of current MLLM-based IQA methods in perceiving fundamental visual features and identify that these limitations stem primarily from suboptimal training strategies.

\item \textbf{Insights into Visual Feature Perception:} Based on a component-wise analysis, we uncover how MLLMs process visual attributes. We highlight the crucial role of the vision extraction module, showing that fine-tuning it induces a notable projection shift in visual tokens as they are aligned with a targeted semantic space.

\end{enumerate}

\section{Related Works}
\label{sec:format}
\begin{figure}[t]
  \centering
  \includegraphics[width=0.5\textwidth]{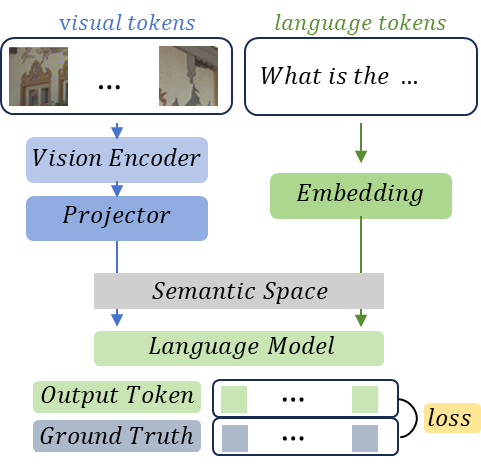}
  \caption{ \textbf{MLLM Structure.} MLLMs integrate three key components: a vision model, a projector model, and a language model. First, visual inputs are processed by the vision encoder, which extracts visual tokens. These tokens are then projected into a shared semantic space using the projector. In this shared space, both the visual features and the language features are jointly processed by the language model, enabling cross-modality understanding. The entire system can be fine-tuned using language model constraints.}
  \label{fig:f2}
\end{figure}

\subsection{Visual Perception in MLLM}
Multi-modal large language models (MLLMs) typically integrate vision and language models to tackle complex real-world tasks. To establish cross-modal understanding, researchers constrain a unified feature space for multi-modal data. Following pioneering MLLM works \cite{flamingo, llava,mplug2}, a common approach projects visual features into the language space, allowing a language model to process both visual and linguistic inputs without additional modules. A representative example is LLaVA \cite{llava}, which comprises three primary modules—a vision encoder, a projector, and a language model—as illustrated in Fig.~\ref{fig:f2}.

Visual perception is one of the most critical capabilities of MLLMs. To enhance it, researchers have explored complex multi-modal patterns \cite{llava, seagull, qground}, such as bounding boxes and local masks. However, it remains uncertain whether MLLMs truly perceive and recognize visual features, and which specific components bolster robust visual perception. In this work, we investigate the learning process of visual perception in MLLMs and seek to elucidate the underlying mechanisms that drive their visual understanding.

\subsection{Image Quality Assessment Using MLLMs} 
Image quality assessment is a foundational task in computer vision. Traditional IQA models often rely on contrastive learning frameworks such as MoCo \cite{moco} and SimCLR \cite{simclr}, classifying distortions and assigning quality scores based on identified impairments. With the advent of MLLMs demonstrating strong image understanding and explanatory capabilities, researchers \cite{qinstruct, seagull, depict} have begun exploring the use of MLLMs as backbones for IQA. In IQA, the primary challenge lies in effectively capturing visual features and deriving quality assessments.

Recent work has sought to teach MLLMs about quality factors through multi-modal quality data. However, images contain significantly more redundant information than language, and some IQA approaches \cite{qinstruct,qalign} appear to over-fit to training templates, sacrificing fundamental visual perception (e.g., recognizing distortions). Although such methods may yield strong performance on numerical quality predictions, their limited perceptual fidelity poses challenges for aligning model outputs with human visual systems. In this study, we investigate the visual perception process in MLLMs, aiming to provide insight into MLLMs’ behavior and to foster broader applications in related downstream tasks.

\section{Methodology}
\label{sec:pagestyle}

\subsection{Overview}
\label{3.1}
We adopt mPLUG-Owl2 \cite{mplug-owl2} as our backbone. As illustrated in Fig.~\ref{fig:f2}, images are first encoded by a vision encoder, and a projector module then maps these features into the token space. Next, the LLM processes the token embeddings for textual reasoning. In this work, we selectively activate specific modules during training to investigate each module’s impact on visual perception learning.

\begin{figure}[!t]
  \centering
  \includegraphics[width=0.5\textwidth]{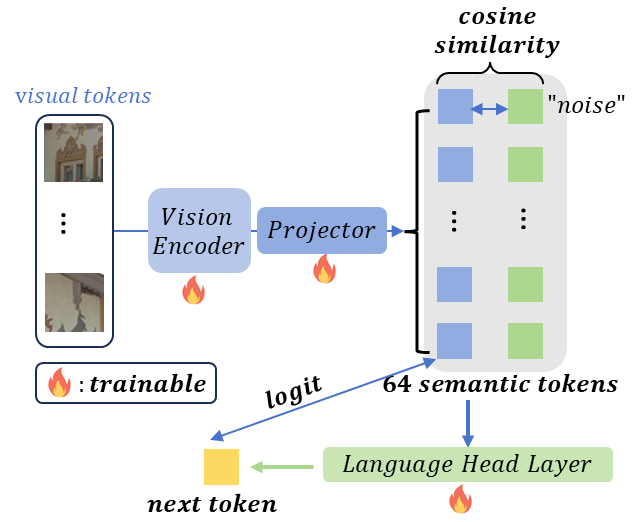}
  \caption{ \textbf{Semantic Distance.} The visual tokens are transferred into semantic space via a vision encoder and projector. There are two direct insights of visual representation. One is comparing the cosine similarity between transferred visual tokens and the language tokens in semantic space. The other is measuring the probability (logit) of the next token.}
  \vspace{-2mm}
  \label{fig:f3}
\end{figure}

\subsection{Data Preparation}
\label{3.2}
\begin{figure}[!h]
    \centering
    \vspace{-2mm}
    \includegraphics[width=\linewidth]{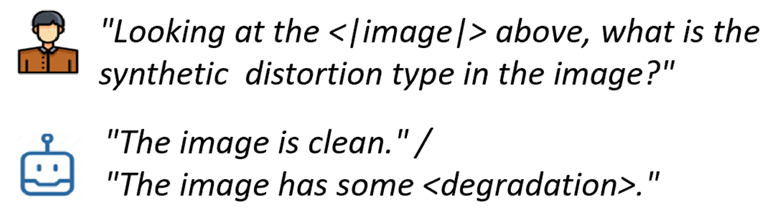}
    \caption{Conversation Template for Distortion Perception.}
    \label{fig:f3.5}
    \vspace{-2mm}
\end{figure}

General IQA datasets consist of original images and synthetic degradations; the degradation labels across the datasets vary a lot. To simplify and unify the degradation types across the several datasets, we utilize the seven most common degradations, including blur, noise, brightness, compression, contrast, colorfulness, and jitter. Additionally, we include a new class called ``clean" for authentic images. Following the general instruction tuning, we create a simple template. The specific degradation type is filled in the conversation template as in Fig. \ref{fig:f3.5}.

\subsection{Visual Perception Investigation}
\label{3.3}
To investigate the visual perception capability of MLLMs, we compare the semantic shifts of visual features before and after training, as illustrated in Fig. \ref{fig:f3}. Two possible changes can occur in the latent visual embeddings. The first involves bringing visual tokens closer to their corresponding labels in the semantic space. In such cases, we expect an increase in the cosine similarity between visual tokens and their corresponding labels, computed using Eq.(\ref{eq1}): \begin{align} \small &{Similarity} = \frac{\cos{({V},L)}-\cos{(V^{init},L)}} {|\cos{(V^{init},L)}|}, \label{eq1} \end{align} \begin{align} \small &{Logit} = \frac{{p(V,L)-p^{init}(V^{init},L)}} {|p^{init}(V^{init},L)|}, \label{eq2} \end{align}
where $cos$ denotes the cosine function, ${V,L}$ are the visual and language embeddings, ``$init$" indicates the naive model, and $p(\cdot)$ is the language model head layer.

The second possible change is an increase in the probability (logit) that the corresponding label will appear as the next token, implying that the visual embeddings better match the label’s context distribution. For example, if people often say ``jpeg" has some ``blur," then, given the prompt ``jpeg has some," a GPT model \cite{gpt} is likely to predict ``blur" as the next token. In this scenario, ``jpeg" has ``some" serves as the context for ``blur," meaning images are embedded in a way that aligns more with ``jpeg" than ``blur." To verify this, we predict the next token based on the visual embeddings, as stated in Eq.~(\ref{eq2}).

\subsection{Fine-tuning}
\label{3.4}
Following the general methodology of MLLMs, we leverage the labeling smooth negative log-likelihood loss function in Eq. \ref{eq4} to the output of the language model (LM). 
\begin{align}
\small
    \mathcal{L}(\boldsymbol{\theta})  
= (1 - \epsilon) \, \cdot
  \Bigl[-\log p_\theta\bigl(y \mid x\bigr)\Bigr]+\epsilon \, \cdot \frac{1}{C}
  \Bigl[\sum_{c=1}^{C} \bigl(-\log p_\theta(c \mid x)\bigr)\Bigr]
\
\label{eq4}
\end{align}
where $\theta$ is the model parameters, $\{x,y\}$ are the input and true label, $C$ is the number of classes and $\epsilon$ is the hyper-parameter.

Technically, we could summarize mPLUG-Owl2 \cite{mplug-owl2} within three parts: the vision part, the projection part, and the LM part. To investigate the learning mechanism, we control the active parameters during the backpropagation separately.

\begin{figure*}[t]
    \centering
    \includegraphics[width=1.00\textwidth]{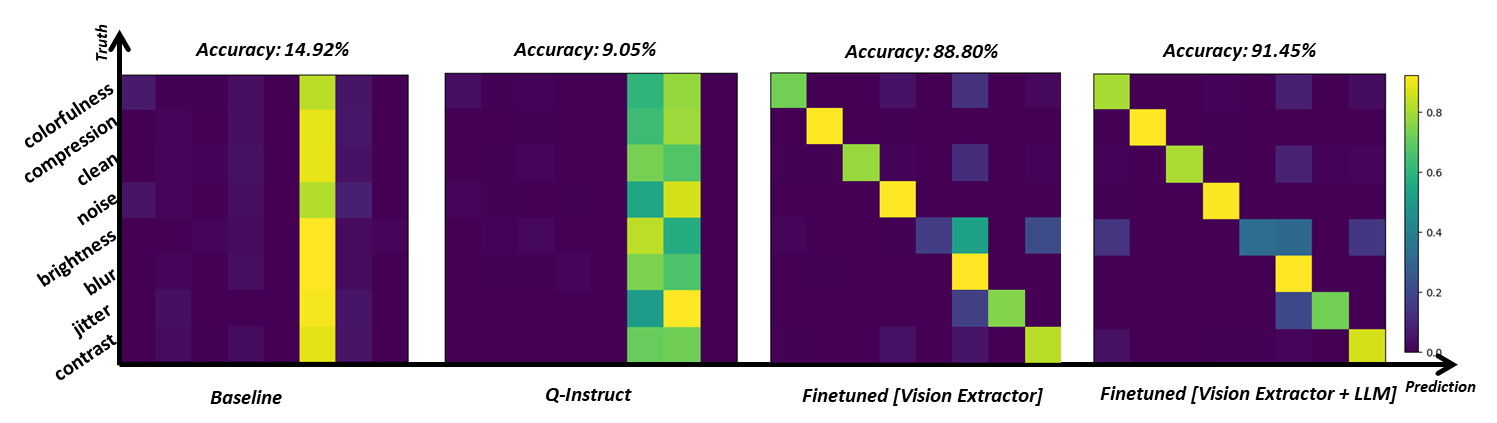}
    \caption{\textbf{Accuracy comparison tested on mixed datasets.} We present the confusion matrix for distortion classification tasks. The Vision Extractor (representing the visual encoder and projector) plays a crucial role. Fine-tuning the Vision Extractor alone significantly improves distortion perception, indicating that LLM reasoning is not the primary component in this case.}
    \label{fig:f4}
\end{figure*}
\section{Experiments}
\label{experiments}
\subsection{Implementation Settings}
\label{4.1}
We use a pre-trained mPLUG-Owl2-7B checkpoint \cite{mplug-owl2} as our backbone. More specifically, the vision encoder is Vit-L/14, and the language model is Llama2-7B \cite{llama2}. The images are resized to \(448 \times 448\) and transferred to 64 semantic tokens.
We construct the four IQA datasets, including LIVE\cite{live}, CSIQ]\cite{csiq}, KADID \cite{kadid}, TID2013 \cite{tid2013}, into image text pairs and randomly split the mix the sets into training and test datasets under the ratio 8:2. The training set totally contains 14306 samples.
The models are fine-tuned on four RTX A6000 GPUs and evaluated on a single RTX A6000. For each training, the model is fine-tuned around 20,000 iterations using AdamW \cite{adamw} optimizer with $\beta1$ = 0.9, $\beta2$ = 0.98, and $\epsilon$ =1e-6.

\begin{table}[!t]
    \centering
    \begin{tabular}{ccc|c}
        \toprule
        \bfseries Encoder & \bfseries Projector & \bfseries LM Layer & \bfseries Accuracy (\%) \\
        \midrule
        \checkmark & - & -          & \bfseries83.43 \\
        - & \checkmark & -          & 72.12 \\
        - & -& \checkmark& 74.24 \\
        \checkmark & \checkmark & - & \bfseries88.80 \\
        \checkmark & - & \checkmark $_{partial}$ &\bfseries 91.45 \\
        - & \checkmark & \checkmark &  74.15 \\
        \bottomrule
        baseline \cite{mplug-owl2} & N.A. & N.A.&14.92\\
        Q-Instruct \cite{qinstruct} & N.A.& N.A. &9.05\\
         \bottomrule
    \end{tabular}
    \caption{\textit{\textbf{Accuracy}}.Fine-tuned modules and their accuracy. In the 5-th experiment, we fine-tuned part of the language model layers.}
    \label{t1}
\end{table}

\subsection{Accuracy Evaluation}
We utilize the mixed test datasets stated in section \ref{4.1}, which contains 3576 images. For our fine-tuned model, we keep the prompt the same as the training phase, that is \textit{``Looking at the $\langle$image$\rangle$ above, what is the synthetic distortion in the image?"}. For baseline \cite{mplug-owl2} model, we add the selections as \textit{``Select from the following options: 1. Blur; 2. Noise; 3. Brightness; 4. Compression; 5. Contrast; 6. Colorfulness; 7. Jitter; 8. Clean"} to the end in order to get a uniform prediction. We plot the accuracy results in Fig. \ref{fig:f4}. The baseline model \cite{mplug-owl2} predominantly classifies all images as blur-related distortions, while another MLLM-based model \cite{qinstruct} heavily overfits to ``blur" and ``jitter," performing at near-random accuracy. After fine-tuning, our model achieves 91.45\% accuracy in distortion recognition. Notably, the vision encoder proves most effective, and even when the language model is frozen, the system learns to generate a response in the format: ``The image has some ⟨degradation⟩." This finding indicates that the vision extractor is the key component in visual perception tasks, and its token embeddings strongly influence the LLM’s reasoning. To delve deeper into why the vision extractor excels, we perform semantic similarity analyses to track how image features and tokens evolve before and after fine-tuning.

\subsection{Semantic Similarity}
\begin{table}[h]
    \centering
    \begin{tabular}{c|c|c|c}
        \toprule
             & \textbf{Projector} & \textbf{Encoder} & \textbf{LM layer} \\
        \midrule
         \textbf{Similarity (\%)} & 87.06 & \textbf{181.0} & 0.42 \\
        \midrule
         \textbf{Logits (\%)} & \textbf{14.51} & 10.96 & 2.04 \\
        \bottomrule
    \end{tabular}
    \caption{\textbf{Semantic shift.} Similarity and logits are relative distances introduced in Section \ref{3.3}. The projector, encoder, and LM layer represent the fine-tuned modules in the model.}
    \label{tab:t2}
\end{table}

\begin{figure}[t]
  \centering
  \includegraphics[width=0.45\textwidth]{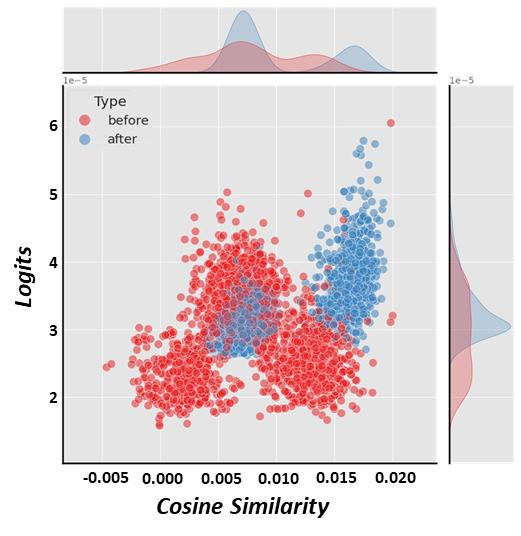}
  \caption{ \textbf{Joint distribution between cosine similarity and logits.} ``before" and ``after" denote before and after fine-tuning. }
  \label{fig:f5}
\end{figure}

n Table \ref{tab:t2}, we measure the semantic shift in MLLM using three models, each fine-tuned by activating only the encoder, projector, or LLM layer. The similarity and logit scores are computed using Eqs.(\ref{eq1}) and (\ref{eq2}), respectively. These findings confirm our hypotheses in Section\ref{3.3}, showing both a reduction in the semantic distance between visual and language tokens and an increase in the logit corresponding to the correct label in next-token prediction. A closer examination reveals that while the encoder primarily reduces the semantic distance between visual and language features, the projector is key to aligning these features with the context distribution.

For a more intuitive understanding of this semantic shift, we randomly select approximately 2,000 samples from the four test datasets described in Section~\ref{4.1} and plot the joint distribution of cosine similarity and logit values before and after fine-tuning. We observe a clear shift toward higher feature alignment in both measures.

\section{Discussion \& Limitation}
\label{sec:majhead}
In this work, we employ a low-level perception task to evaluate the learning and perception capabilities of multi-modal large language models. Our experiments demonstrate that while MMLMs can perceive and reason about fundamental visual features, their behavior often degrades due to overfitting. This issue can be effectively mitigated by fine-tuning the vision encoder, which significantly reduces the semantic gap between visual and linguistic representations, thereby improving low-level distortion recognition. Our findings highlight the importance of dedicated training constraints on the vision extraction module for generating text-explainable features and achieving coherent reasoning in vision-centric tasks.

However, our study has several limitations. First, we do not compare our approach with other vision-language models, such as LLaVA \cite{llava}, which we plan to explore in future work. Second, our experiments are conducted on a selected set of distortion types and IQA datasets, potentially limiting the generalizability of our findings to other forms of visual features or more complex quality assessment scenarios. Finally, while our component-wise analysis offers meaningful insights into model behavior, further investigation is needed to explore alternative architectures or training paradigms that could mitigate overfitting, improve generalization, and foster more coherent and interpretable reasoning in vision-centric tasks.

\vfill\pagebreak

% -------------------------------------------------------------------------
{
\bibliographystyle{IEEEbib}
\bibliography{strings,refs}
}
\end{document}